\documentclass[letterpaper, 10 pt, conference]{ieeeconf}

\IEEEoverridecommandlockouts %

\usepackage{amsmath,amsfonts,amssymb}
\usepackage{amsfonts} %
\usepackage{mathrsfs} %
\usepackage{pgfplots}

\usepackage{algorithm}
\usepackage{algpseudocode}

\usepackage[tight]{subfigure}
\usepackage{tabularx}
\usepackage{multirow}
\usepackage{siunitx}

\usepackage[export]{adjustbox}

\usepackage{caption}
\usepackage{graphicx}
\graphicspath{{figures/}}
\usepackage{import}

\usepackage{xargs} %

\usepackage[style=ieee, backend=biber, bibencoding=utf8, maxbibnames=5, mincitenames=1, maxcitenames=2, url=false, doi=false, isbn=false, citestyle=numeric-comp, natbib=true, eprint=false]{biblatex}

\usepackage[colorinlistoftodos,prependcaption,textsize=small]{todonotes}

\newcommandx{\change}[2][1=]{\todo[inline,linecolor=blue,backgroundcolor=blue!25,bordercolor=blue,#1]{#2}}

\usepackage{bm} %
\usepackage{lipsum}
\newcommand{\Secref}[1]{Section~\ref{#1}} 

\usepackage{scalerel} %

\newcommand \Global {\scaleobj{0.95}{\square}}
\newcommand \Next       {\scaleobj{0.85}{\bigcirc}}
\newcommand \Final       {\lozenge}
\newcommand \lUntil      {\mathbin{\mathcal{U}\kern-.1em}} %
\newcommand \ImpliesCustomized {\scaleobj{0.9}{\implies}}

\newcommand\copyrighttext{%
	\scriptsize \textcolor{blue}{\textcopyright 2019 IEEE. Personal use of this material is permitted.  Permission from IEEE must be obtained for all other uses, in any current or future media, including reprinting/republishing this material for advertising or promotional purposes, creating new collective works, for resale or redistribution to servers or lists, or reuse of any copyrighted component of this work in other works}}
\newcommand\copyrightnotice{%
	\begin{tikzpicture}[remember picture,overlay]
	\node[anchor=north,yshift=-7.5pt] at (current page.north) {\fbox{\parbox{\dimexpr\textwidth-\fboxsep-\fboxrule\relax}{\copyrighttext}}};
	\end{tikzpicture}%
}

\pgfplotsset{compat=newest}
\pgfplotsset{plot coordinates/math parser=false}
\usetikzlibrary{decorations.markings, positioning}

\usepackage{booktabs}

\newlength\figureheight 
\newlength\figurewidth 

\title{\LARGE \bf From Specifications to Behavior: Maneuver Verification in a Semantic State Space}
\author{Klemens Esterle$^{1}$, Vincent Aravantinos$^{1}$ and Alois Knoll$^{2}$%
	\thanks{$^{1}$Klemens Esterle and Vincent Aravantinos are with fortiss GmbH, An-Institut Technische Universit\"{a}t M\"{u}nchen, Munich, Germany}%
	\thanks{$^{2}$Alois Knoll is with Robotics, Artificial Intelligence and Real-time Systems, Technische Universit\"{a}t M\"{u}nchen, Munich, Germany}%
}

\begin{document}

\maketitle
\copyrightnotice
\thispagestyle{empty}
\pagestyle{empty}

\global\csname @topnum\endcsname 0
\global\csname @botnum\endcsname 0

\newcommand{\figurename}{Fig. }

\newcommand {\vect} {\boldsymbol}
\newcommand {\matr} {\boldsymbol}

\newcommand{\state} {\vect{x}}
\newcommand{\stateSpace} {\vect{\mathcal{X}}}
\newcommand{\beliefstate} {\vect{b}}
\newcommand{\contr} {\vect{u}}
\newcommand{\contrSpace} {\vect{\mathcal{U}}}
\newcommand{\meas} {\vect{y}}
\newcommand{\procNoise}{\vect{w}}

\newcommand {\cov}  {\matr{\Sigma}}

\newcommand{\stateNoDelta}{\hat\state}
\newcommand{\contrNoDelta}{\hat\contr}
\newcommand{\procNoiseNoDelta}{\hat\procNoise}

\newcommand{\abc}[2][\empty]{%
  \ifthenelse{\equal{#1}{\empty}}
    {no opt, mand.: \textbf{#2}}
    {opt: \textbf{#1}, mand.: \textbf{#2}}
}

\newcommand {\noiseu} {\procNoise}
\newcommand {\covu} {\matr{\Sigma_{\noiseu,}}}

\newcommand {\noiseuNoDelta} {\procNoiseNoDelta}
\newcommand {\covuNoDelta} {\matr{\Sigma_{\noiseuNoDelta,}}}

\newcommand {\defnoiseu}[1][\empty]{
 \ifthenelse{\equal{#1}{\empty}}
    {\noiseu\sim N(0,\covu)}
    {\noiseu_{#1}\sim N(0,\covu_{#1})}
}

\newcommand {\defnoiseuNoDelta}[1][\empty]{
 \ifthenelse{\equal{#1}{\empty}}
    {\noiseuNoDelta\sim N(0,\covuNoDelta)}
    {\noiseuNoDelta_{#1}\sim N(0,\covuNoDelta_{#1})}
}

\newcommand {\covm} {\matr{R}}
\newcommand {\noisem} {\vect{\nu}}
\newcommand {\defnoisem}[1][\empty]{
 \ifthenelse{\equal{#1}{\empty}}
    {\noisem\sim N(0,\covm)}
    {\noisem_{#1}\sim N(0,\covm_{#1})}
}

\newcommand{\stateB}{\vect{\xi}}
\newcommand{\contrB}{\vect{\nu}}
\newcommand{\procNoiseB}{\vect{\omega}}

\newcommand{\AB}{\mathcal{A}}
\newcommand{\BB}{\mathcal{B}}
\newcommand{\WB}{\mathcal{W}}
\newcommand{\costStateB}{\mathcal{Q}}
\newcommand{\costContrB}{\mathcal{R}}
\newcommand{\covStatesB}{\mathcal{S}_{\state}}
\newcommand{\covProcNoiseB}{\mathcal{S}_{\procNoise}}

\newcommand{\FB}{\mathcal{F}}

\newcommand{\cct}{\vect{t}}
\newcommand{\ccsval}{s}
\newcommand{\ccT}{\matr{T}}
\newcommand{\ccsvec}{\vect{s}}

\newcommand{\costState}{\matr{Q}}
\newcommand{\costContr}{\matr{R}}
\newcommand{\feedbackMatrix}{\matr{K}}
\newcommand{\cost}{J}

\newcommand{\stateConstraintMatrix}{\matr{C}}
\newcommand{\stateConstraintVector}{\vect{c}}

\newcommand{\stateConstraintFunc}{c}

\newcommand{\contrConstraintMatrix}{\matr{D}}
\newcommand{\contrConstraintVector}{\vect{d}}

\newcommand{\contrConstraintFunc}{d}

\newcommand{\stateRef}{\state^{*}}
\newcommand{\contrRef}{\contr^{*}}

\newcommand{\stateDelta}{\Delta\state}
\newcommand{\contrDelta}{\Delta\contr}

\newcommand {\partialder}[4][\bigg]{\frac{\partial #2}{\partial #3}#1|_{#4}}
\newcommand {\partialdernoarg}[3][\bigg]{\frac{\partial #2}{\partial #3}#1}

\newcommand{\nat}{\mathbb{N}}
\newcommand{\real}{\mathbb{R}}
\newcommand{\compl}{\mathbb{C}}

\newcommand{\norm}[1]{\left\| #1 \right\|}

\newcommand{\half}{\frac{1}{2}}

\newcommand{\parenth}[1]{ \left( #1 \right) }
\newcommand{\bracket}[1]{ \left[ #1 \right] }
\newcommand{\accolade}[1]{ \left\{ #1 \right\} }
\newcommand{\pardevS}[2]{ \delta_{#1} f(#2) }
\newcommand{\pardevF}[2]{ \frac{\partial #1}{\partial #2} }

\newcommand{\vecii}[2]{\begin{pmatrix} #1 \\ #2 \end{pmatrix}}
\newcommand{\veciii}[3]{\begin{pmatrix}  #1 \\ #2 \\ #3	\end{pmatrix} }
\newcommand{\veciv}[4]{\begin{pmatrix}  #1 \\ #2 \\ #3 \\ #4	\end{pmatrix}}

\newcommand{\matii}[4]{\left[ \begin{array}[h]{cc} #1 & #2 \\ #3 & #4 \end{array} \right]}
\newcommand{\matiii}[9]{\left[ \begin{array}[h]{ccc} #1 & #2 & #3 \\ #4 & #5 & #6 \\ #7 & #8 & #9	\end{array} \right]}

\newcommand{\transp}{^{\intercal}}
\newcommand{\Reg}{$^{\textregistered}$}
\newcommand{\reg}{$^{\textregistered}$ }
\newcommand{\Tm}{\texttrademark}
\newcommand{\tm}{\texttrademark~}
\newcommand {\bsl} {$\backslash$}

\newtheorem{theorem}{Theorem}[section]
\newtheorem{lemma}[theorem]{Lemma}
\newtheorem{corollary}[theorem]{Corollary}
\newtheorem{remark}[theorem]{Remark}
\newtheorem{definition}[theorem]{Definition}
\newtheorem{equat}[theorem]{Equation}
\newtheorem{example}[theorem]{Example}
\newcommand{\insertfigure}[4]{ %
	\begin{figure}[htbp]
		\begin{center}
			\includegraphics[width=#4\textwidth]{#1}
		\end{center}
		\vspace{-0.4cm}
		\caption{#2}
		\label{#3}
	\end{figure}
}

\newcommand{\refFigure}[1]{\figurename \ref{#1}}
\newcommand{\refChapter}[1]{chapter \ref{#1}}
\newcommand{\refSection}[1]{section \ref{#1}}
\newcommand{\refParagraph}[1]{paragraph \ref{#1}}
\newcommand{\refEquation}[1]{(\ref{#1})}
\newcommand{\refTable}[1]{Table \ref{#1}}
\newcommand{\refAlgorithm}[1]{Algorithm \ref{#1}}

\newcommand{\rigidTransform}[2]
{
	${}^{#2}\!\mathbf{H}_{#1}$
}

\newcommand{\code}[1]
 {\texttt{#1}}

\newcommand{\comment}[1]{\marginpar{\raggedright \noindent \footnotesize {\sl #1} }}

\newcommand{\clearemptydoublepage}{%
  \ifthenelse{\boolean{@twoside}}{\newpage{\pagestyle{empty}\cleardoublepage}}%
  {\clearpage}}

\newcommand{\etAl}{\emph{et al.}\mbox{ }}

\newcommand{\todoi}[1]{\todo[inline]{#1}}

\begin{abstract}
To realize a market entry of autonomous vehicles in the foreseeable future, the behavior planning system will need to abide by the same rules that humans follow.
Product liability cannot be enforced without a proper solution to the approval trap.
In this paper, we define a semantic abstraction of the continuous space and formalize traffic rules in linear temporal logic (LTL). Sequences in the semantic state space represent maneuvers a high-level planner could choose to execute. We check these maneuvers against the formalized traffic rules using runtime verification.
By using the standard model checker NuSMV, we demonstrate the effectiveness of our approach and provide runtime properties for the maneuver verification.
We show that high-level behavior can be verified in a semantic state space to fulfill a set of formalized rules, which could serve as a step towards safety of the intended functionality.
\end{abstract}

\IEEEpeerreviewmaketitle

\section{Introduction}
\label{sec:introduction}

\subsection{Motivation}
\label{subsec:motivation}
A lot of effort has been put into demonstrating the feasibility of behavior planning, but an approach guaranteeing safety of the intended function (SOTIF, see \cite{IOS2019}) is still far away. Building up a safety case for an autonomous driving system based on mileage driven is not suitable due to the high-dimensional state space of real-world scenarios. Instead, advances in verifiability, safety assessment and explainability will be essential to make autonomous driving reliable \cite{Schwarting2018}.

Traffic rules have been designed to help humans manage the otherwise chaotic traffic environment. If all vehicles were fully autonomous, the current set of rules could be reduced or adapted. In a transition period with mixed traffic, however, we need to make sure that the planned behavior satisfies those specified rules at all times.
First of all, obeying those rules would make the behavior of autonomous cars more understandable to other traffic participants. Secondly, it will help clarify the liability of the autonomous vehicle in case of an accident.

For the behavior of an autonomous vehicle to follow traffic rules, large state-machines have been used (e.g. \cite{Urmson2008, Montemerlo2008, Leonard2008}), requiring extensive and careful engineering. Transferring such state machines to different markets with other specifications does not scale well and quickly becomes hard to maintain. 
The problem of behavior planning is challenging due to the high-dimensional continuous state space, non-linear motion dynamics, interactions with other agents and their unobservable intentions.
This has led to researchers investigating collaborative approaches \cite{Kessler2017a}, game-theoretic approaches \cite{Lenz2016}, as well as probabilistic approaches \cite{Hubmann2018}. 

However, since these new approaches try to incorporate reactive predictions into the decision making process, the complexity is growing and it becomes even harder to fuse them with classic rule-based systems to abide by the traffic regulations and specifications. Rules get either hard-coded or are inherently learned from data, which are not transferable solutions in case of rule changes.
Specification as the satisfaction of traffic rules should thus be formalized in some way, for instance as formal logic. 

\begin{figure}[tb]
\footnotesize
\centering
\def\svgwidth{\columnwidth}
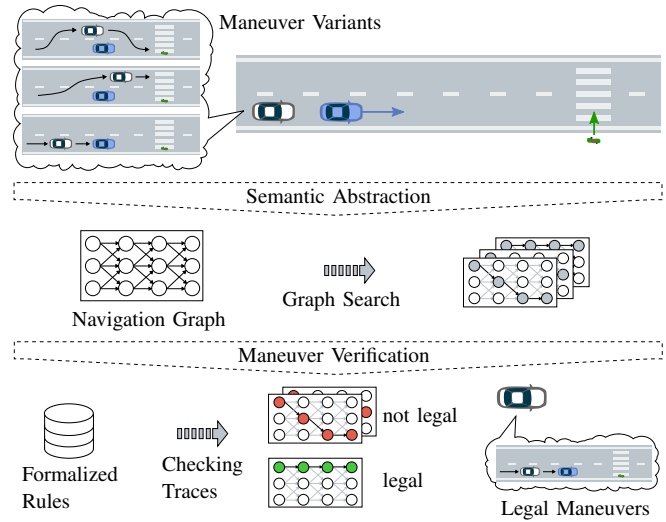
\caption{Overview of our approach. The combinatorial maneuver variants are made accessible by building up a semantic graph structure of the predicted evolution of the scenario. A graph search yields maneuvers, which we check against formalized rules.}
\label{fig:intro}
\end{figure}

\subsection{Contribution}
Checking potentially many trajectory hypotheses for complex traffic rules in a continuous space entails high computational costs. In this paper, we address the problem of verifying the high-level behavior of an autonomous vehicle in a semantic state space to satisfy a set of traffic rules. The solution shall be easily integrated into a planning framework so that the final trajectory also satisfies the rules and does not need to be checked again. We make use of the work of \cite{Altche2018a}, building our verification approach on top of that. \refFigure{fig:intro} gives an overview of our approach.

The main contributions of this paper are:
\begin{itemize}
	\item an LTL representation corresponding to the partitioning method of \cite{Altche2018a},
	\item the formalization of some traffic rules to LTL for the introduced semantic state,
	\item a method to verify high-level maneuvers to comply to these rules,
	\item evaluations showing computational properties of the approach.
\end{itemize}

This paper is organized as follows: \Secref{sec:related_work} presents work directly related to our paper. \Secref{sec:problem} defines the addressed problem. \Secref{sec:preliminaries} introduces the necessary concepts this work is based on. The proposed method is presented in \Secref{sec:method}. In \Secref{sec:rules_translation}, we formalize a set of traffic rules. \Secref{sec:evaluation} evaluates the algorithm's abilities followed by a discussion in \Secref{sec:conclusion}.

\section{Related Work}
\label{sec:related_work}
Recent work such as \cite{Wolff2014, Rizaldi2018, Vasile2017} investigated verifying reach-and-avoid tasks\footnote{In reach-and-avoid tasks, the robot stays in a safe region until the goal region is reached.} or routing tasks in linear temporal logic (LTL). LTL is a formal language that combines boolean and temporal operators. \citet{Plaku2016} argue for its expressiveness on sequencing and partial ordering as well as persistency (\textit{"move slowly"}). \citet{Wolff2014} transform linear temporal logic constraints to mixed integer linear constraints. In this way, temporal constraints can be integrated into optimization-based motion planning without creating an abstraction to verify the LTL formula. Solving this MILP leads to an optimal trajectory satisfying the logical specifications. They apply this approach to reach-avoid tasks, whereas we focus on maneuver planning.

\citet{Rizaldi2018} present an automata-based maneuver planner where each motion primitive is encoded as a state. They formalize reach ("\textit{until you reach the goal}") and avoid tasks ("\textit{do not collide with an obstacle}") to LTL to check the validity of a transition from one state to another. In contrast, we identify atomic propositions that allow us to express high-level maneuvers such as overtaking directly in LTL.

\citeauthor{Rizaldi2015} were the first to use logic as a language to formalize Vienna Road Traffic rules by specifically using higher order logic (HOL) in \cite{Rizaldi2015}. Their work is extended in \cite{Rizaldi2017}, where they use LTL to formalize German traffic rules. As they use high-level states such as \textit{overtaking} or \textit{merging} as atomic propositions, they need to verify the value of the \textit{overtaking} proposition externally (i.e., not LTL). In contrast, we describe maneuvers such as overtaking using LTL as a sequence of states. Their approach checks for rule satisfaction in a continuous state space and can be used as the monitor system of an ego trajectory, whereas we formulate rules in a semantic state space.

\citet{Kohlhaas2015} use a semantic state space that consists of the state and action space of the ego vehicle for a high-level maneuver planner. Relations of objects and road elements are represented in an ontology. They transform the semantic space as a directed graph and use a graph search to generate feasible solutions. They claim that their semantic state space is capable of representing traffic rules by the use of conditions. However, they do not describe which rules they model. In contrast, we present a set of rules from the Vienna Convention in regard to overtaking and passing orders.

\citet{ReyesCastro2013} translate LTL specifications into an automaton. A second automaton is created as a discrete abstraction of an RRT*-explored workspace. A discrete search then creates feasible traces satisfying the LTL formula which guide the RRT* into feasible directions satisfying the specification. Similar to our approach, they check for the satisfaction of rules at an abstract layer. However, they only check for safety rules such as \textit{"do not cross solid center lines"} or \textit{"do not travel in the wrong direction"}. In contrast, the combinatorial approach we use directly applies lane constraints to the planning problem, thus satisfying these types of rules by design.

\section{Problem Definition}
\label{sec:problem}
We follow the behavior planning problem formulation we introduced in \cite{Esterle2018}. There, we separate the problem into high-level envelope planning and low-level trajectory planning inside a homotopy. Based on an object list and a prediction for each object, the envelope planner has to plan a maneuver envelope

\begin{figure}[tb]
\vspace{0.15cm}
\footnotesize
\centering
\def\svgwidth{\columnwidth}
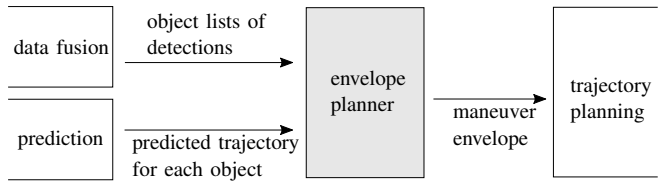
\caption{The envelope planner unifies maneuver generation and verification. The semantic representation is calculated from an object list and a predicted trajectory of each object. Our contribution is the highlighted block.}
\label{fig:contribution}
\end{figure}

\begin{equation} 
\bm{\zeta}_{p} = [s_{max}, \> s_{min}, \> d_{left}(s_i), \> d_{right}(s_i)]
\label{equ:maneuver_envelope}
\end{equation}
at each prediction time step $\theta_p$ in local street-wise coordinates with the arc length $s$ and the perpendicular offset $d$. The maneuver envelope is then used for low-level trajectory planning. \refFigure{fig:contribution} shows our contribution embedded in a planning framework.

In this paper, our goal is to generate envelopes satisfying a set of rules. To do so, we generate a semantic trace $\tau$ representing a possible maneuver variant. When selecting such a maneuver, we need to consider rules such as passing orders and overtaking rules. We formulate the satisfaction of traffic regulations and social conventions on the high-level envelope planner as a model checking problem.
Given a semantic trace $\tau$ and a set of rules $R_i$, the following should hold:
\begin{equation}
\tau \models \bigwedge_{i}^{n} R_i.
\label{equ:model_checking_problem}
\end{equation}

The operator $\models$ denotes the satisfaction relation of LTL.

\section{Preliminaries}
\label{sec:preliminaries}

\subsection{Linear Temporal Logic}
\label{subsec:ltl}

Temporal logics extend classical logics by temporal operators to reason not just about an absolute truth but about truths which might hold only at some points in time.
Linear temporal logic (LTL) is one of the simplest, building upon propositional logic by adding temporal operators.

\subsubsection{Syntax}
Formally, the language $\rho$ of LTL formulas is defined as
\begin{equation}
	\rho ::= A \, | \, \lnot A \, | \, \rho_1 \land \rho_2 \, | \, \rho_1 \lor \rho_2 \, | \rho_1 {\mathbin{\kern-.3em\ImpliesCustomized\kern-.3em}} \rho_2 \, | \, \Next \rho \, | \, \rho_1 \lUntil \rho_2 \, | \, \Global \rho \, | \, \Final \rho,
	\nonumber
\end{equation}
where $A$ denotes a set of \emph{atomic} propositions, $\lnot$ (resp. $\land$, $\lor$, $\ImpliesCustomized$) denote the boolean operators ``not'', ``and'', ``or'' and ``implies'',
and $\Next$, (resp. $\lUntil$, $\Global$, $\Final$) denote the temporal operators ``next'', ``until'', ``globally'' (or ``always''), ``finally'' (or ``eventually''). 

Precedence is defined as follows: The unary operators bind stronger than the binary ones, e.g. $\rho_1 \lUntil \Next \rho_2$ is equal to \\$\rho_1 \lUntil (\Next \rho_2)$. The ``until'' operator is right-associative, e.g. $\rho_1 \lUntil \rho_2 \lUntil \rho_3$ is equal to $\rho_1 \lUntil (\rho_2 \lUntil \rho_3)$. $\lnot$ and $\Next$ bind equally strong. Operator $\lUntil$ takes precedence over $\land$, $\lor$, $\ImpliesCustomized$.

We first need to characterize when a formula is true or not.
In LTL, this is done with respect to a \emph{trace}: 
\begin{definition}
  A \emph{valuation} of $A$ is a function that assigns to each proposition of $A$ an element of $\{\mathbf T, \mathbf F\}$.
  A \emph{trace} is an infinite sequence of valuations.
\end{definition}
For instance, if $A=\{X,Y,Z\}$, then the function $f$ defined by $f(X)=\mathbf{T}$, $f(Y)=\mathbf{F}$, $f(Z)=\mathbf{T}$ is a valuation of $A$.
Intuitively, it represents the truth value of every variable in a given configuration.
A trace represents the evolution of such configurations over time. Note that from an algorithmic viewpoint, a maneuver defined over a given planning horizon is finite. A finite sequence of valuations is turned into a trace by repeating the last valuation indefinitely.
\subsubsection{Semantics}
We can define the semantics of an LTL formula as follows:
\begin{definition}
  For every formula $\rho$, every instant $i$ and every trace $\tau$,
  the relation $\tau,i\models \rho$ (pronounced ``$\rho$ holds for $\tau$ at instant $i$'') is inductively defined as follows:
  \begin{itemize}
    \item $\tau,i\models a\in A$ if and only if (iff) $\tau_i=\mathbf T$,
    \item $\tau,i\models \rho_1\land\rho_2$ iff $\tau,i\models \rho_1$ and $\tau,i\models\rho_2$,
    \item $\tau,i\models \rho_1\lor\rho_2$ iff $\tau,i\models \rho_1$ or $\tau,i\models\rho_2$,
    \item $\tau,i\models \rho_1\ImpliesCustomized\rho_2$ iff $\tau,i\not\models \rho_1$ or $\tau,i\models\rho_2$,
    \item $\tau,i\models \Next\rho$ iff $\tau,i+1\models \rho$,
    \item $\tau,i\models \Global\rho$ iff for all $j\geq i$, $\tau,j\models \rho$,
    \item $\tau,i\models \Final\rho$ iff there exists $j\geq i$ s.t. $\tau,j\models \rho$,
    \item $\tau,i\models \rho_1\lUntil\rho_2$ iff there exists $j\geq i$ such that $\tau,j\models \rho_2$ and for all $k$ $i\leq k\leq j$, $\tau,k\models\rho_1$.
  \end{itemize}
\end{definition}
When we have a trace $\tau$ with the configurations $s_i$ with $i \in {1,2,3}$, we denote it by ($s_1\rightarrow s_2 \rightarrow s_3$).
\refTable{tab:ltl_formula} provides examples for the temporal operators.
\begin{table}[tb]
	\vspace{0.15cm}
	\caption{Truth values of LTL formulas $\rho$ are displayed in a receding fashion for the atomic propositions $A=\{x,y\}$ at time instants $i=1,...,4$.}
	\begin{displaymath}  %
	  \footnotesize
	  \begin{array}{ccccccc}\toprule
	    \rho & A & i=1 & i=2 & i=3 & i=4 & ... \\
	    \cmidrule{3-7}
      	&x & \mathbf T & \mathbf F & \mathbf T & \mathbf T & \mathbf T \\ 
	    &y & \mathbf F & \mathbf T & \mathbf F & \mathbf F & \mathbf F \\ 
	    \midrule
      	\Next x & & \mathbf F & \mathbf T & \mathbf T & \mathbf T & \mathbf T \\
	    \Global x & & \mathbf F & \mathbf F & \mathbf T & \mathbf T & \mathbf T \\
	    \Final y & & \mathbf T & \mathbf T & \mathbf F & \mathbf F & \mathbf F \\
	    y \lUntil x & & \mathbf T & \mathbf T & \mathbf T & \mathbf T & \mathbf T \\
	    \bottomrule
	  \end{array}
	\end{displaymath}
	\label{tab:ltl_formula}
\end{table}

\subsection{Free Space-Time Partitioning}
\label{subsec:partitioning}
To facilitate high-level decisions, an intermediate representation bridging the gap between obstacle lists from sensor fusion and semantic information is needed. We build on top of the work of \citet{Altche2018a}, that partitions collision-free space-time into discrete three-dimensional cells with geometric adjacency relations, described in the following.

\begin{figure}[tb]
\footnotesize
\centering
\def\svgwidth{0.8\columnwidth}
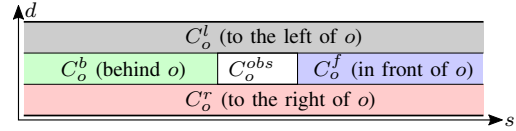
\caption{Partitioning of the two-dimensional space around a single obstacle $o$ into four collision-free regions $C_o^i$., from \cite{Altche2018a}.}
\label{fig:freespacetimepartitioning}
\end{figure}

\begin{figure}[tb]
\footnotesize
\centering
\def\svgwidth{\columnwidth}
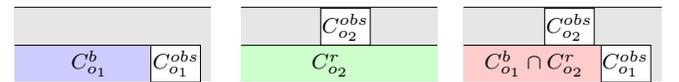
\caption{Set-based union of two regions $C_{o_1}^b$ and $C_{o_2}^r$ to form a region that describes the free space "to the right of $o_1$ and behind $o_2$".}
\label{fig:scheme_partitioning_o1_o2}
\end{figure}

We first briefly introduce the concept of partitioning in 2D space. Let $\mathcal{Q}$ denote the extend of the road in the Frenet space. The free space $\mathcal{Q}_f$ around an obstacle $o$ is partitioned into four collision-free regions -- front, behind, left and right. For each obstacle, regions are derived in accordance with \refFigure{fig:freespacetimepartitioning}.
\refFigure{fig:scheme_partitioning_o1_o2} illustrates the concept of enumeration of the free space. The set-based intersection of the region "behind $o_1$" (denoted by $C^b_{o_1}$) and "to the right of $o_2$" ($C^r_{o_2}$) yields a new region "to the right of $o_1$ and behind $o_2$". This region is abbreviated with the signature $\sigma=(rb)$, where the order of the letters refers to the respective obstacle $o_i$. 

Partitioning the 2D free space at each discrete time step yields a set of free space-time cells $E_{\sigma}^p$, denoted by $\mathcal{P}$. $E_{\sigma}^p$ denotes the space-time cell corresponding to the signature $\sigma$ at time step $\theta_p$. The term \textit{cell} is used here in contrast to 2D, where we only speak of \textit{regions}. 

From this discrete cell decomposition, adjacency information is calculated to create a graph structure. The discrete transition graph $\mathcal{G}$ is defined as $\mathcal{G} = (\mathcal{V}, \mathcal{E})$ with the vertex set $\mathcal{V}$ and the edges set $\mathcal{E}$.
We let $\mathcal{V} = \mathcal{P}$. Edges between two cells $E_{\sigma_1}^p$ and $E_{\sigma_2}^{p+1}$ at two consecutive time steps $\theta_p$ and $\theta_{p+1}$ exist if the cells are adjacent at both time steps. For the definition of adjacency $\text{adj}_{\theta_p}(\cdot,\cdot)$, we refer the reader to \cite{Altche2018a}.
Formally, $\mathcal{E}$ is defined as: 
\begin{equation}
\mathcal{E}=\big\{(E_{\sigma_1}^p, E_{\sigma_2}^{p+1}) | E_{\sigma_1}^p, E_{\sigma_2}^{p+1} \in \mathcal{V}, \text{adj}_{\theta_p} (\sigma_1, \sigma_2) = 1\big\}.
\nonumber
\end{equation}

\begin{figure}[tb]
\vspace{0.18cm}
\footnotesize
\centering
\def\svgwidth{\columnwidth}
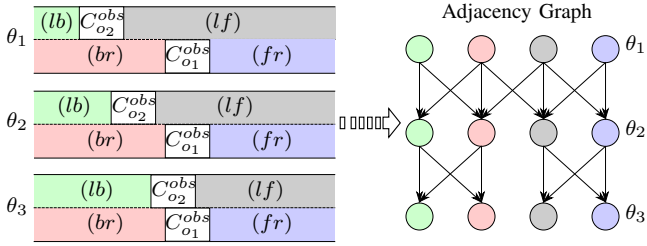
\caption{Mapping of an obstacle-related partitioning of the free space to a directed graph, that incorporates the adjacency information of the free space-time partitions.}
\label{fig:discrete_partitioning_graph_example}
\end{figure}

\refFigure{fig:discrete_partitioning_graph_example} illustrates the relationship between the free space-time partitioning and the adjacency graph, called \textit{navigation graph}. Since the obstacle $o_2$ moves faster than $o_1$, the cells with the signatures $(br)$ and $(lf)$ are no longer adjacent at $\theta_3$. Therefore, no connection exists in the adjacency graph between the corresponding vertices.

Each path $(E_{\sigma_1}^0, ..., E_{\sigma_{m+1}}^{m+1})$ in $\mathcal{G}$ corresponds to a set of homotopic collision-free trajectories (that may, however, be infeasible). This reduces the behavior generation problem to a discrete selection of a maneuver variant and trajectory planning for the obtained maneuver variant. 

\section{Maneuver Verification}
\label{sec:method}

\subsection{Semantic State Space Definition}
\label{subsec:ap_definition}
The challenge of formalizing traffic rules in LTL starts with the selection of an expressive semantic state space and abstracting that as atomic propositions.
Based on the navigation graph, we introduce a semantic state space for each agent in the scenario except the ego vehicle:
\begin{itemize}
	\item Obstacle-related:
	\begin{itemize}
		\item position of the free space in relation to the traffic participant (in front $f$, behind $b$, to the left $l$, to the right $r$),
		\item traffic participant type vehicle $v$, pedestrian $p$, cyclist $c$, rail-borne $r$ (denoted as a subscript of the position).
	\end{itemize}
	\item Road type-related (pedestrian crosswalk $\mathcal{R}_{pc}$, carriageway $\mathcal{R}_{cw}$).
\end{itemize}
Let us formalize these notions in a form suitable for LTL.
We first need to define atomic propositions.
There exist only two road type propositions: $\mathcal{R}_{pc}$ and $\mathcal{R}_{cw}$.
For obstacle propositions, let $v_1,...,v_{n_v}$ (resp. $p_1,...,p_{n_p}$, resp. $c_1,...,c_{n_c}$) be the set of all vehicles (resp. pedestrians, resp. cyclists) in the scenario.
Then, for every obstacle $o\in\{v_1,...,v_{n_v},p_1,...,p_{n_p},c_1,...,c_{n_c}\}$, the following are atomic propositions: $f_o$, $b_o$, $l_o$, $r_o$.

Let $\alpha$ be a representation of the current situation.
We can create a corresponding LTL trace, in the following denoted by $\tau_{\alpha}$ (recall that a trace is a sequence of functions assigning to every atomic variable a truth value at a given instant).
First, we have $\tau_{\alpha_i}(\mathcal{R}_{pc}) = \mathbf T$ (resp. $\tau_{\alpha_i}(\mathcal{R}_{cw}) = \mathbf T$)
iff the road type is a pedestrian crosswalk (resp. on the carriageway) at instant $i$.

For every instant $i$ and every object $o$, we have:
\begin{itemize}
  \item $\tau_{\alpha_i}(f_o) = \mathbf T$ iff ego is in front of $o$ at instant $i$,
  \item $\tau_{\alpha_i}(b_o) = \mathbf T$ iff ego is behind $o$ at instant $i$,
  \item $\tau_{\alpha_i}(l_o) = \mathbf T$ iff ego is to the left of $o$ at instant $i$,
  \item $\tau_{\alpha_i}(r_o) = \mathbf T$ iff ego is to the right of $o$ at instant $i$.
\end{itemize}

\subsection{Road type-based Partitioning}
\label{subsec:roadtype-partitioning}
Following our state space definition, we extend the partitioning of the road space $\mathcal{Q}$ (see \Secref{subsec:partitioning}) to include road type information. Instead of checking the road-type for each free space-time cell when evaluating a trace, we choose to encode the property of road-type in the discrete graph. This way, we keep the direct mapping from a trace to a maneuver intact instead of introducing an intermediate step of processing a discrete path and cutting out road-type related regions.

We calculate a set of road regions $\mathcal{P}^r$, where the superscript $r$ denotes the partitioning related to the road type. We then use $\mathcal{P}^r$ as the initial set of regions for the free space-time partitioning, following Algorithm 1 in \cite{Altche2018a}.

\subsection{Runtime Verification}
\label{subsec:online_verification}
Classical model checking would require translating the graph structure into an automaton. The model checker would then generate traces that comply with the formalized rules. However, we choose to go in another direction by performing runtime verification. We do not create an automaton from the full graph but only from a graph traversal from the root node to a goal node thus yielding only a trace instead of a "full-fledged" automaton. We then use a model checker to check whether such a trace $\tau$ satisfies the specified rules according to \refEquation{equ:model_checking_problem}. The benefit of our formulation is that we leave the general concept of \cite{Altche2018a, Park2015} untouched, which is to assign heuristic costs to each graph solution, then rank them along those costs and select a finite number of graph solutions for trajectory planning. Using additional information as graph costs such as time gaps would not directly be possible if we translated the graph to an automaton for model checking. Another benefit of runtime verification is its possible use in combination with other planners as the trace to be checked does not necessarily have to be a graph solution. 

However, the creation of a single automaton for each trace instead of the whole graph creates additional computational overhead. Note that we do not know beforehand whether the selected path from the transition graph is dynamically feasible. It thus may be necessary to check many traces. The model checking approach on the other hand would always provide valid plans.

\begin{algorithm}[b]
	\caption{Maneuver Verification on Navigation Graph}
	\label{algo:maneuver_verification}
	\begin{algorithmic}[1]
		\State \textbf{Input} $\Delta\theta$: planning interval
		\State \textbf{Input} $n$: number of time steps
		\State \textbf{Input} env: environment
		\State \textbf{Output} $T_{sat}$: set of rule-satisfying traces
		\State $T_{sat}$ $\gets$ $\emptyset$
		\State $\mathcal{P} \gets$ generateFiniteAbstraction(env, $\Delta\theta$, $n$)\;
		\State $\mathcal{G}$ $\gets$ generateAdjacencyGraph($\mathcal{P}$, env)
		\State $\sigma_{ego}^{\theta_{0}}$ $\gets$ getRootVertice($\mathcal{G}$, env)
		\State $\mathcal{V}^{\theta_{n-1}}$ $\gets$ getGoalVerticeCandidates($\mathcal{G}$, env)
		\For{\textbf{each} $\sigma^{\theta_{n-1}}$ in $\mathcal{V}^{\theta_{n-1}}$}
		\State $\tau \gets$ findAllTraces($\mathcal{G}$, $\sigma_{ego}^{\theta_{0}}$, $\sigma^{\theta_{n-1}}$)
		\State costs$[\tau] \gets$ calculateTimeMargins($\tau$)
		\EndFor
		\State $T_{sorted}\gets$ sort($T$, costs)
		\While{$T_{sorted} \neq \emptyset$}
		\If {satisfiesLTLSpec($\tau$)}
		\State $T_{sat} \gets T_{sat} \lUntil \tau$
		\EndIf
		\EndWhile
	\end{algorithmic}
\end{algorithm}

\refAlgorithm{algo:maneuver_verification} shows the proposed method. For a given planning interval $\Delta\theta$, we construct the semantic finite abstraction of the current state of all relevant traffic participants and their predicted motions by first partitioning the state space (Line 6). The function generateFiniteAbstraction() contains the partitioning along the road type (\Secref{subsec:roadtype-partitioning}) and the free space (see \Secref{subsec:partitioning}). For a detailed description of the free space partitioning and the calculation of adjacency in generateAdjacencyGraph(), we refer the reader to \cite{Altche2018a}. From that, we obtain a graph representation (Line 7). The starting node $\sigma_{ego}^{\theta_{0}}$ represents the cell where the ego vehicle is located at the initial planning time step. As we do not know the final cell of the optimal maneuver yet, all cells at the final time step $\mathcal{V}^{\theta_{n-1}}$ are possible goal cells. We perform a graph search to get all possible traces $T$ (Line 11). We finally calculate the time margins using a function calculateTimeMargins() as in \cite{Altche2018a}. We assume the existence of a function sort() to sort the traces by their costs (Line 14). We then verify each semantic trace $\tau$ using the proposed verification method.

We turn a finite time trace into an infinite trace by repeating the last state indefinitely. This may cause inaccuracies, as we evaluate the next-operator on the last (repeated) state. Investigating these side-effects in detail and evaluating finite-time semantics is the subject of future work.

As the instants $i$ correlate with prediction time steps, a verified trace from $T_{sat}$ satisfying the specifications can be directly mapped to a maneuver envelope as defined in \refEquation{equ:maneuver_envelope}, serving as the input to the trajectory planner.

\section{Vienna Convention Traffic Rules as LTL}
\label{sec:rules_translation}

\newtheorem{vienna-rule}{Traffic Rule}
Based on the Vienna Convention of Traffic Rules \cite{EconomicCommissionforEuropeInlandTransportCommittee1968}, we have selected rules that cover the interaction of the ego vehicle with only one other traffic participant. Extending this to rules that cover more traffic participants is the subject of future work.
\begin{vienna-rule}
Do not overtake a vehicle on its right side, except in congested traffic, where overtaking on the right is also allowed \textit{(Article 11.1, 11.6)}.
\end{vienna-rule}
The rule concerns roads with two or more lanes in the same direction such as highways. The rule shall only apply in congested traffic. Formalizing congestion requires the speed of the vehicles, which we choose not to include in our semantic state space definition. Here, we assume the scene interpretation module to provide a signal $\texttt{CONGESTED}$. It could for example be set to true if all other vehicles are slower than a certain threshold. We formulate $R_1$ as:
\begin{equation}
R_1(v): \lnot\texttt{CONGESTED} \ImpliesCustomized \Global \lnot (b_v \land \Next (b_v \lUntil r_v \lUntil f_v)).
\nonumber
\end{equation}

\begin{figure}[tb]
\vspace{0.15cm}
\footnotesize
\centering
\def\svgwidth{0.89\columnwidth}
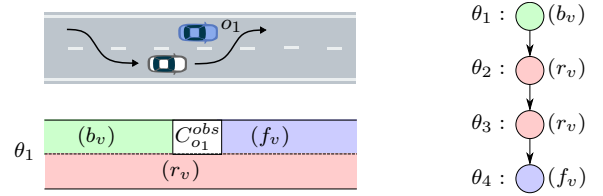
\caption{Traffic Rule 1: Possible maneuver overtaking the red vehicle $o_1$ on the right. The partitioning around $o_1$ at $\theta_1$ is displayed, which stays similar for the time steps $\theta_{1..4}$. The semantic trace $(b_v \rightarrow r_v \rightarrow r_v \rightarrow f_v)$ of that maneuver violates $R_1$.}
\label{fig:rule1_explaination}
\end{figure}

This means that except for the case of congested traffic, we do not allow a temporal sequence in which the ego vehicle starts behind a vehicle, goes to the right of the vehicle and then goes in front of the vehicle. The until operator allows to follow this rule independently of the number of instants that the ego vehicle stays to the right of the other vehicle. 
If the first occurrence of $b_v$ were not in the formula, the trace $(r_v \rightarrow f_v)$ would not be valid. We thus limit the forbidden set by adding $b_v$ followed by the ordering $b_v \lUntil r_v \lUntil f_v$. 
\refFigure{fig:rule1_explaination} illustrates a scenario in which the obtained trace for the overtaking maneuver on the right does violate $R_1$. Note that since the rule applies to each obstacle, we perform the rule checking for each obstacle separately. For simplicity reasons, \refFigure{fig:rule1_explaination} displays only an abbreviated signature of the trace omitting road type information.

\begin{vienna-rule}
Do not overtake a vehicle that slows down or waits in front of a pedestrian walk \textit{(Article 11.9)}.
\end{vienna-rule}

\begin{figure}[tb]
\footnotesize
\centering
\def\svgwidth{\columnwidth}
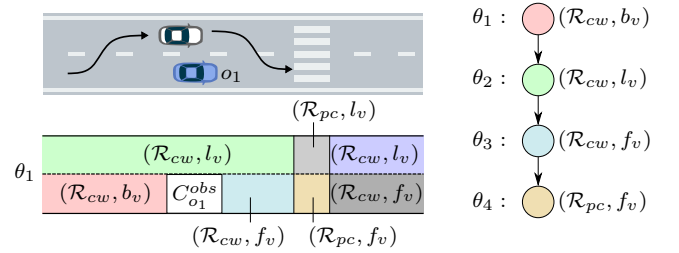
\caption{Traffic Rule 2: The red vehicle approaches a pedestrian walk. A possible maneuver overtaking the red vehicle $o_1$ on the left is illustrated. The partitioning around $o_1$ at $\theta_1$ is displayed. The semantic trace $(\mathcal{R}_{cw},b_v \rightarrow \mathcal{R}_{cw},l_v \rightarrow \mathcal{R}_{cw},f_v \rightarrow \mathcal{R}_{pc},f_v)$ of that maneuver violates $R_2$.}
\label{fig:rule2_explaination}
\end{figure}

We rephrase as follows: \textit{A vehicle shall not be overtaken if a pedestrian crosswalk is immediately in front of the vehicle}. This is a slightly modified version of the actual rule as we do not model the slow-down but only the position of the crosswalk in relation to the vehicle. This way we impose that a vehicle which is not slowing down should not be overtaken as well, which we consider a reasonable assumption as well. Formally, this results in

\begin{equation}
R_2(v): \Global \lnot (b_v \land \Next (b_v \lUntil l_v \lUntil (f_v \land \mathcal{R}_{pc}))).
\nonumber
\end{equation}
The rule is similar to $R_1$, mirroring the overtaking maneuver on the right instead of the left. The last temporal state we request for such a trace to be forbidden needs to be in front of the respective vehicle but also of the road type \textit{pedestrian crosswalk} $\mathcal{R}_{pc}$. \refFigure{fig:rule2_explaination} illustrates a scenario where the maneuver to overtake the red vehicle right in front of a pedestrian crosswalk will be forbidden as it violates $R_2$.

\begin{figure}[tb]
\vspace{0.15cm}
\footnotesize
\centering
\def\svgwidth{\columnwidth}
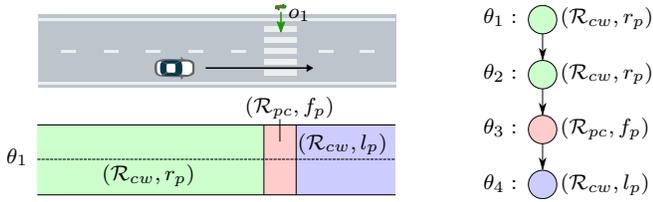
\caption{Traffic Rule 3: A pedestrian to the left of the ego vehicle intends to use a pedestrian crossing. A possible maneuver to pass before the pedestrian crosses the street is illustrated. The semantic trace $(\mathcal{R}_{cw},r_p \rightarrow \mathcal{R}_{cw},r_p \rightarrow \mathcal{R}_{pc},f_p \rightarrow \mathcal{R}_{cw},l_p)$ of that maneuver violates $R_3$.}
\label{fig:rule3_explaination}
\end{figure}

\begin{vienna-rule}
Stop if pedestrians are using or intend to use a pedestrian crossing. \textit{(Article 21.3)}
\end{vienna-rule}
We formulate it as
\begin{equation}
R_3(p): \Global \; \lnot (\mathcal{R}_{pc} \land f_p),
\nonumber
\end{equation}
such that no state in the trace is allowed that is a pedestrian walk and lies in front of a pedestrian. The formulation covers pedestrians coming from the left and right. \refFigure{fig:rule3_explaination} illustrates a maneuver that $R_3$ forbids.

\section{Evaluation}
\label{sec:evaluation}
\subsection{Evaluating the Formalized Rules}
\label{subsec:eval_formalized}
We will first use synthetic traces to show the effectiveness of our formalized rules. \refTable{tab:ltl_satisfaction_r1} shows that the trace $\tau_5$ $(b_v \rightarrow r_v \rightarrow r_v \rightarrow f_v)$ representing a maneuver to overtake on the right  violates $R_1$. It can also be seen that the sequence defining a maneuver to overtake on the right $(b_v \rightarrow r_v \rightarrow f_v)$ in the trace $\tau_8$ can be captured and classified as violating as well, no matter how many states the trace has.
\begin{table}[tb]
\caption{Example traces $\tau_i$ and whether they satisfy rule $R_1$. We omit the road type information $\mathcal{R}$ in the LTL trace as it does not have any impact on the satisfaction of $R_1$.}
\begin{displaymath}
\begin{array}{c|c c c c c c | c}
\toprule
\text{trace} \; \tau_i & \theta_0 & \theta_1 & \theta_2 & \theta_3 & \theta_4 & \theta_5 & \tau_i \models R_1\\
\midrule
\tau_1& b_v & b_v & l_v & f_v &  &  & \mathbf{T} \\
\tau_2& b_v & l_v & l_v & b_v &  &  & \mathbf{T} \\
\tau_3& b_v & b_v & r_v & b_v &  &  & \mathbf{T} \\
\tau_4& r_v & r_v & f_v & f_v &  &  & \mathbf{T} \\
\tau_5& b_v & r_v & r_v & f_v &  &  & \mathbf{F} \\
\tau_6& b_v & r_v & f_v & f_v &  &  & \mathbf{F} \\
\tau_7& b_v & r_v & f_v & r_v &  &  & \mathbf{F} \\
\tau_8& b_v & r_v & r_v & b_v & r_v & f_v & \mathbf{F} \\
\bottomrule
\end{array}
\end{displaymath}

\label{tab:ltl_satisfaction_r1}
\end{table}
However, there are some drawbacks from the selected state space. The relation "right of $o_i$" states that $o_i$ is in the next lane to the right. Still, we currently cannot capture that $o_i$ could be $30\si{\meter}$ before or behind the ego vehicle. An extension of the state space could be beneficial to formalize such rules more clearly. We leave this to future work. 

Rule $R_2$ forbids overtaking of a vehicle that is right in front of a pedestrian walk. \refTable{tab:ltl_satisfaction_r2} shows example traces and whether they satisfy rule $R_2$ or not. In regard to the obstacle-relation $(b_v \rightarrow b_v \rightarrow l_v \rightarrow f_v)$, $\tau_{i|1..4}$ and $\tau_5$ are identical. However, they differ in the road type of their final state, which is why $\tau_5$ violates $R_2$ and the others do not.

\begin{table}[tb]
\vspace{0.15cm}
\caption{Example traces $\tau_i$ and whether they satisfy rule $R_2$.}
\begin{displaymath}
\begin{array}{c|c c c c | c}
\toprule
\text{trace} \; \tau_i & \theta_0 & \theta_1 & \theta_2 & \theta_3 & \tau_i \models R_2\\
\midrule
\tau_1& R_{cw}; b_v & R_{cw}; b_v & R_{cw}; l_v & R_{cw}; f_v & \mathbf{T} \\
\tau_2& R_{cw}; b_v & R_{pc}; b_v & R_{cw}; l_v & R_{cw}; f_v & \mathbf{T} \\
\tau_3& R_{cw}; b_v & R_{cw}; b_v & R_{cw}; l_v & R_{pc}; f_v & \mathbf{F} \\
\bottomrule
\end{array}
\end{displaymath}

\label{tab:ltl_satisfaction_r2}
\vspace{-0.2cm} %
\end{table}

Rule $R_3$ forbids any state that is a pedestrian crosswalk and in front of a pedestrian. \refTable{tab:ltl_satisfaction_r3} shows a trace $\tau_1$ with ($r_p \rightarrow f_p \rightarrow f_p \rightarrow l_p$). This represents a situation in which a pedestrian is to the left of the street oriented towards the street, while the ego vehicle passes the pedestrian. As can be seen in \refTable{tab:ltl_satisfaction_r3}, rule $R_3$ forbids such a trace if the pedestrian stands at a pedestrian crosswalk, represented by the state $\mathcal{R}_{pc},f_p$. If not, the trace will be classified as legal.
\begin{table}[tb]
\caption{Example traces $\tau_i$ and whether they satisfy rule $R_3$}
\begin{displaymath}
\begin{array}{c|c c c c | c}
\toprule
\text{trace} \; \tau_i & \theta_0 & \theta_1 & \theta_2 & \theta_3 & \tau_i \models R_3\\
\midrule
\tau_1& R_{cw}; r_p & R_{cw}; f_p & R_{cw}; f_p & R_{pc}; l_p & \mathbf{T} \\
\tau_2& R_{cw}; l_p & R_{cw}; f_p & R_{cw}; f_p & R_{pc}; r_p & \mathbf{T} \\
\tau_3& R_{cw}; l_p & R_{pc}; f_p & R_{pc}; f_p & R_{cw}; r_p & \mathbf{F} \\
\bottomrule
\end{array}
\end{displaymath}

\label{tab:ltl_satisfaction_r3}
\end{table}

\subsection{Maneuver Verification}
We want to answer the following questions:
\begin{itemize}
	\item Is it feasible to check all maneuvers?
	\item How long does it take to check a maneuver?
	\item How does the complexity of the problem influence the runtime?
\end{itemize}
\subsubsection{Implementation} As explained in \Secref{subsec:online_verification}, we do runtime verification instead of model checking. Although it might appear as an overkill, we use NuSMV, a powerful model checking tool \cite{Cimatti2002}, since this will allow us to easily modify and extend our approach in the future. All algorithms have been implemented in MATLAB. The simulations were performed on a 2.6GHz desktop PC with 16GB RAM.
\subsubsection{Scenarios} 
\begin{figure}[tb]
\vspace{0.15cm}
\centering
\footnotesize
\subfigure{\label{subfig:a}\def\svgwidth{\columnwidth}
\begingroup%
  \makeatletter%
  \providecommand\color[2][]{%
    \errmessage{(Inkscape) Color is used for the text in Inkscape, but the package 'color.sty' is not loaded}%
    \renewcommand\color[2][]{}%
  }%
  \providecommand\transparent[1]{%
    \errmessage{(Inkscape) Transparency is used (non-zero) for the text in Inkscape, but the package 'transparent.sty' is not loaded}%
    \renewcommand\transparent[1]{}%
  }%
  \providecommand\rotatebox[2]{#2}%
  \newcommand*\fsize{\dimexpr\f@size pt\relax}%
  \newcommand*\lineheight[1]{\fontsize{\fsize}{#1\fsize}\selectfont}%
  \ifx\svgwidth\undefined%
    \setlength{\unitlength}{327.88417306bp}%
    \ifx\svgscale\undefined%
      \relax%
    \else%
      \setlength{\unitlength}{\unitlength * \real{\svgscale}}%
    \fi%
  \else%
    \setlength{\unitlength}{\svgwidth}%
  \fi%
  \global\let\svgwidth\undefined%
  \global\let\svgscale\undefined%
  \makeatother%
  \begin{picture}(1,0.05581239)%
    \lineheight{1}%
    \setlength\tabcolsep{0pt}%
    \put(0,0){\includegraphics[width=\unitlength,page=1]{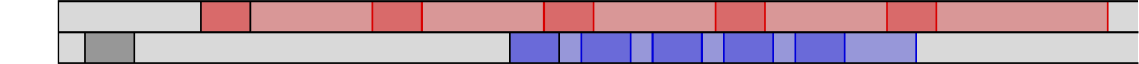}}%
    \put(0.09595549,0.03570945){\color[rgb]{0,0,0}\makebox(0,0)[lt]{\lineheight{1.25}\smash{\begin{tabular}[t]{l}ego\end{tabular}}}}%
    \put(0,0){\includegraphics[width=\unitlength,page=2]{Scenario1_discription.pdf}}%
    \put(0.40046783,0.00662053){\color[rgb]{0,0,0}\makebox(0,0)[lt]{\lineheight{1.25}\smash{\begin{tabular}[t]{l}$o_2$\end{tabular}}}}%
    \put(0,0){\includegraphics[width=\unitlength,page=3]{Scenario1_discription.pdf}}%
    \put(0.20229284,0.00662053){\color[rgb]{0,0,0}\makebox(0,0)[lt]{\lineheight{1.25}\smash{\begin{tabular}[t]{l}$o_1$\end{tabular}}}}%
    \put(0,0){\includegraphics[width=\unitlength,page=4]{Scenario1_discription.pdf}}%
    \put(-0.0020253,0.02169961){\color[rgb]{0,0,0}\makebox(0,0)[lt]{\lineheight{1.25}\smash{\begin{tabular}[t]{l}$\mathcal{S}_1$\end{tabular}}}}%
  \end{picture}%
\endgroup%
}
\subfigure{\label{subfig:b}\def\svgwidth{\columnwidth}
\begingroup%
  \makeatletter%
  \providecommand\color[2][]{%
    \errmessage{(Inkscape) Color is used for the text in Inkscape, but the package 'color.sty' is not loaded}%
    \renewcommand\color[2][]{}%
  }%
  \providecommand\transparent[1]{%
    \errmessage{(Inkscape) Transparency is used (non-zero) for the text in Inkscape, but the package 'transparent.sty' is not loaded}%
    \renewcommand\transparent[1]{}%
  }%
  \providecommand\rotatebox[2]{#2}%
  \newcommand*\fsize{\dimexpr\f@size pt\relax}%
  \newcommand*\lineheight[1]{\fontsize{\fsize}{#1\fsize}\selectfont}%
  \ifx\svgwidth\undefined%
    \setlength{\unitlength}{327.75949553bp}%
    \ifx\svgscale\undefined%
      \relax%
    \else%
      \setlength{\unitlength}{\unitlength * \real{\svgscale}}%
    \fi%
  \else%
    \setlength{\unitlength}{\svgwidth}%
  \fi%
  \global\let\svgwidth\undefined%
  \global\let\svgscale\undefined%
  \makeatother%
  \begin{picture}(1,0.09244584)%
    \lineheight{1}%
    \setlength\tabcolsep{0pt}%
    \put(0,0){\includegraphics[width=\unitlength,page=1]{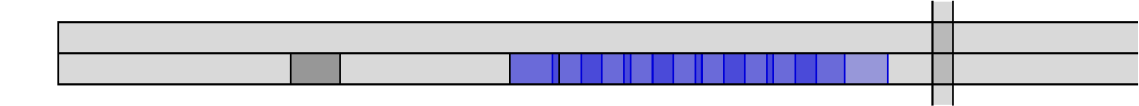}}%
    \put(0.28362972,0.05425448){\color[rgb]{0,0,0}\makebox(0,0)[lt]{\lineheight{1.25}\smash{\begin{tabular}[t]{l}ego\end{tabular}}}}%
    \put(0,0){\includegraphics[width=\unitlength,page=2]{Scenario2_discription.pdf}}%
    \put(0.40062016,0.02515449){\color[rgb]{0,0,0}\makebox(0,0)[lt]{\lineheight{1.25}\smash{\begin{tabular}[t]{l}$o_1$\end{tabular}}}}%
    \put(0,0){\includegraphics[width=\unitlength,page=3]{Scenario2_discription.pdf}}%
    \put(-0.00202607,0.04023931){\color[rgb]{0,0,0}\makebox(0,0)[lt]{\lineheight{1.25}\smash{\begin{tabular}[t]{l}$\mathcal{S}_2$\end{tabular}}}}%
  \end{picture}%
\endgroup%
}
\subfigure{\label{subfig:c}\def\svgwidth{\columnwidth}
\begingroup%
  \makeatletter%
  \providecommand\color[2][]{%
    \errmessage{(Inkscape) Color is used for the text in Inkscape, but the package 'color.sty' is not loaded}%
    \renewcommand\color[2][]{}%
  }%
  \providecommand\transparent[1]{%
    \errmessage{(Inkscape) Transparency is used (non-zero) for the text in Inkscape, but the package 'transparent.sty' is not loaded}%
    \renewcommand\transparent[1]{}%
  }%
  \providecommand\rotatebox[2]{#2}%
  \newcommand*\fsize{\dimexpr\f@size pt\relax}%
  \newcommand*\lineheight[1]{\fontsize{\fsize}{#1\fsize}\selectfont}%
  \ifx\svgwidth\undefined%
    \setlength{\unitlength}{327.75951716bp}%
    \ifx\svgscale\undefined%
      \relax%
    \else%
      \setlength{\unitlength}{\unitlength * \real{\svgscale}}%
    \fi%
  \else%
    \setlength{\unitlength}{\svgwidth}%
  \fi%
  \global\let\svgwidth\undefined%
  \global\let\svgscale\undefined%
  \makeatother%
  \begin{picture}(1,0.14938708)%
    \lineheight{1}%
    \setlength\tabcolsep{0pt}%
    \put(0,0){\includegraphics[width=\unitlength,page=1]{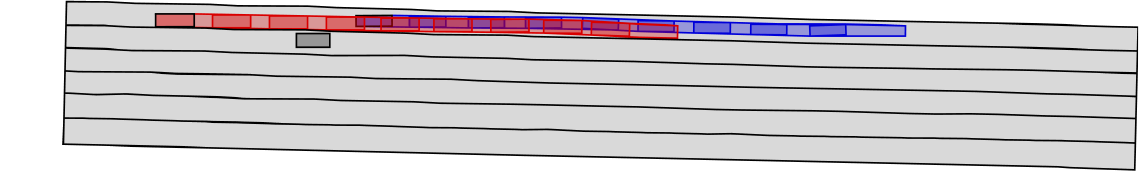}}%
    \put(0.27332088,0.06505873){\color[rgb]{0,0,0}\makebox(0,0)[lt]{\lineheight{1.25}\smash{\begin{tabular}[t]{l}ego\end{tabular}}}}%
    \put(0,0){\includegraphics[width=\unitlength,page=2]{Scenario3_discription.pdf}}%
    \put(0.13256645,0.09018637){\color[rgb]{0,0,0}\makebox(0,0)[lt]{\lineheight{1.25}\smash{\begin{tabular}[t]{l}$o_1$\end{tabular}}}}%
    \put(0,0){\includegraphics[width=\unitlength,page=3]{Scenario3_discription.pdf}}%
    \put(-0.00202607,0.07887443){\color[rgb]{0,0,0}\makebox(0,0)[lt]{\lineheight{1.25}\smash{\begin{tabular}[t]{l}$\mathcal{S}_3$\end{tabular}}}}%
    \put(0.32091781,0.08456443){\color[rgb]{0,0,0}\makebox(0,0)[lt]{\lineheight{1.25}\smash{\begin{tabular}[t]{l}$o_2$\end{tabular}}}}%
    \put(0,0){\includegraphics[width=\unitlength,page=4]{Scenario3_discription.pdf}}%
  \end{picture}%
\endgroup%
}\caption{Scenarios $\mathcal{S}_1$, $\mathcal{S}_2$ and $\mathcal{S}_3$.}
\label{fig:scenarios_under_test}
\end{figure}

We evaluate three different scenarios: A two-lane same-direction scenario similar to \cite{Altche2018a}, a two-lane scenario with a pedestrian crossing, and a three-lane highway \textit{CommonRoad} scenario \cite{Koschi2017}. We assume full observability and use a constant-acceleration model to predict the other participants. \refFigure{fig:scenarios_under_test} illustrates the scenarios. We compare all scenarios for a planning horizon of $4\si{\second}$ for two different grades of discretization. We compute the predicted occupancy of each obstacle between consecutive time instants $[t_k, t_{k+1}]$. This ensures that we do not miss any occupancy when partitioning at larger step sizes. However, the occupied areas grow with the size of the step size. Some maneuvers might thus not be available at a broader discretization.

\subsubsection{Computational Results} \refTable{tab:table_computations} summarizes our computational results. The size of the adjacency graph depends on the number of time steps, the number of obstacles, the number of unique road types and the layout of the scenario.

At this early stage, we use a naive implementation calculating every possible path from the starting node to all possible goal nodes. This leads to an explosion of the number of traces for $\mathcal{S}_2$. To understand the reason for that, let us have a look at the following traces: 

\begin{equation*}
\begin{aligned}
\tau_1: & (b_v \rightarrow b_v \rightarrow l_v) \\
\tau_2: & (b_v \rightarrow l_v \rightarrow l_v).
\end{aligned}
\end{equation*}
The graph approach distinguishes between the traces $\tau_{1,2}$ although they might (and in this case do) represent the same homotopy class. To introduce a topology grouping step after the graph search omitting those duplicate traces or to use branch-and-bound techniques will be the subject of future work. 

The number of traces for $\mathcal{S}_2$ is much higher than for $\mathcal{S}_1$. Because $\mathcal{S}_2$ contains a pedestrian walk, we obtain four additional cells. We get a high number of adjacent cells leading up to a high number of possible traces for two reasons: First, the adjacency definition in \cite{Altche2018a} lets two cells be adjacent if the closures intersect at a single point, which trades a lower number of infeasible traces in a kinematic sense for lower computational costs in the adjacency calculation. Secondly, we do not distinguish between forward and backward driving, which also introduces additional connections between cells. Improving these two points in combination with the topology grouping step will reduce the number of traces.
We list the computational costs for calculating a single trace using the Dijkstra algorithm, for which we set the weights of the graph to the time gaps introduced in \cite{Altche2018a}. Improving the weights of the graph will also be the subject of future work. 

We observe for $\mathcal{S}_1$ with the smaller step size that a small number of traces get rejected. These traces represent maneuvers going to the left lane and then starting an overtaking maneuver on the right which consequently violate $R_1$. Running $\mathcal{S}_1$ with the coarser discretization does not allow such complex maneuvers, which is why all of the 16 traces are valid.

We chose Scenario $\mathcal{S}_2$ to demonstrate the necessity for our rule-checking approach. Roughly forty percent of the traces get rejected, because they violate the rules. Checking a single trace takes around $22\si{\milli\second}$ per obstacle in our implementation. This could be further sped up as we currently create a model file for every trace that we check, save that on the hard drive and then call the model checker. Future work will focus on reducing the computational time.

Scenario $\mathcal{S}_3$, which depicts a scene from the popular NGSIM dataset, shall show the feasibility of our approach in more realistic scenarios. Compared to $\mathcal{S}_1$, it takes roughly three times as much time. Reasons for that are the arbitrary shape of the road and the number of lanes.

The verification of a trace $\tau$ for $\mathcal{S}_2$ takes about half the time compared to $\mathcal{S}_1$ and $\mathcal{S}_3$. The reason is that $\mathcal{S}_1$ and $\mathcal{S}_3$ consider two obstacles, which brings the necessity to check the rules for each obstacle in $\tau$ separately. We evaluated the runtime of a trace verification with ten rules instead of three, which did not change the runtime significantly.
With the need to check potentially many traces, the model checker might seem like a bottleneck. However, model checking tools support a huge set of formulas. To tune the verification algorithm for online use in a car running at a high frequency, a model checker only supporting our limited formula could be implemented more efficiently.

We have not discussed sequential planning yet, which means the use of the maneuver verification approach in a receding horizon manner. A maneuver to overtake a vehicle on the right might be forbidden, but a maneuver to only change to the right lane might be legal. At the next planning step, passing the vehicle and changing back to the left lane might be feasible as well. Possible ways to cope with that could be to incorporate past states ${s_{-1}, s_{-2}, ...}$ into the trace checking. A detailed analysis of such effects is the subject of future work.

\begin{table}[tb]
\vspace{0.15cm}
\centering
\caption{Computation Results for Maneuver Verification.}
\begin{tabular}{@{}lll|ll|ll@{}}
\toprule
& \multicolumn{2}{c}{\textit{$\mathcal{S}_1$}}  & \multicolumn{2}{c}{\textit{$\mathcal{S}_2$}} & \multicolumn{2}{c}{\textit{$\mathcal{S}_3$}}\\
\midrule
& \multicolumn{6}{c}{\textit{Scenario Description}} \\
\cmidrule{2-7}
Planning horizon & 4$\si{\second}$ & 4$\si{\second}$ & 4$\si{\second}$ & 4$\si{\second}$ & 4$\si{\second}$ & 4$\si{\second}$\\
Step size & 1$\si{\second}$ & 0.5$\si{\second}$ & 1$\si{\second}$ & 0.5$\si{\second}$ & 1$\si{\second}$ & 0.5$\si{\second}$\\
No. of obstacles & 2 & 2 & 1 & 1 & 2 & 2\\
No. of road types & 1 & 1 & 2 & 2 & 1 & 1\\
\midrule
& \multicolumn{6}{c}{\textit{Semantic Partitioning}}\\
\cmidrule{2-7}
Graph Nodes & 20 & 36 & 35 & 63 & 20 & 36\\
Graph Edges & 32 & 70 & 124 & 248 & 40 & 80\\
\midrule
& \multicolumn{6}{c}{\textit{Maneuver Verification}}\\
\cmidrule{2-7}
No. of all traces & 16 & 672 & 139 & 75441 & 76 & 4240\\
No. of sat. traces & 16 & 640 & 83 & 35531 & 61 & 1840\\
\midrule
& \multicolumn{6}{c}{\textit{Computation times} $[\si{\second}]$}\\
\cmidrule{2-7}
Partitioning & 0.073 & 0.129 & 0.144 & 0.206 & 0.290 & 0.426\\
Graph generation & 0.014 & 0.021 & 0.020 & 0.037 & 0.013 & 0.020\\
Dijkstra Search & 0.001 & 0.000 & 0.000 & 0.000 & 0.001 & 0.000\\
Calculating costs & 1e-05 & 9e-06 & 1e-05 & 1e-05 & 8e-06 & 1e-05\\
Verification of $\pi^s$ & 0.044 & 0.045 & 0.022 & 0.024 & 0.043 & 0.046\\
\bottomrule
\end{tabular}

\label{tab:table_computations}
\end{table}

\section{Conclusion and Future Work}
\label{sec:conclusion}
To formalize traffic rules using LTL, we introduced a novel link between combinatorial behavior planning and model checking.
We showed that LTL can indeed be used to verify high-level maneuver plans. This may help shift part of the development from writing code to writing specifications.
Although we discretize the continuous environment and evaluate the rules on a higher, more abstract level, the rules still hold on the continuous level due to the homotopy property of each trace.

We selected rules that cover only one other traffic participant. Future work will need to consider interactions with more than one road user. Further investigations will also concentrate on the improvement of the graph costs, where we currently use the notion of time gaps, a relatively sparse definition that yields many traces with the same costs. We briefly touched the limited expressiveness of the semantic space in \Secref{subsec:eval_formalized}, which we will try to extend in the future while carefully weighing increased computational complexity against expressiveness.

When expressing traffic rules in the real world, there is a difference between "should" and "must". Some rules must never be violated, others should not be violated except in emergency situations. Future work includes investigating the difference between soft and hard rules as well as developing a heuristic for a graph search algorithm that incorporates knowledge of previously checked traces.

As a probabilistic prediction will be essential for robust behavior planning, the approach needs to be extended to incorporate the prediction of multiple possible maneuvers. This could open the door for probabilistic model checkers. Alternatively, our method could be used as an additional safety layer which verifies a probabilistic behavior planner.

\renewcommand{\bibfont}{\small}
\printbibliography
\end{document}